# Efficient Probabilistic Inference with Partial Ranking Queries

**Jonathan Huang**
School of Computer Science
Carnegie Mellon University
Pittsburgh, PA 15213

**Ashish Kapoor**
Microsoft Research
1 Microsoft Way
Redmond, WA 98052

**Carlos Guestrin**
School of Computer Science
Carnegie Mellon University
Pittsburgh, PA 15213

## Abstract

Distributions over rankings are used to model data in various settings such as preference analysis and political elections. The factorial size of the space of rankings, however, typically forces one to make structural assumptions, such as smoothness, sparsity, or probabilistic independence about these underlying distributions. We approach the modeling problem from the computational principle that one should make structural assumptions which allow for efficient calculation of typical probabilistic queries. For ranking models, "typical" queries predominantly take the form of partial ranking queries (e.g., given a user's top-$k$ favorite movies, what are his preferences over remaining movies?). In this paper, we argue that riffled independence factorizations proposed in recent literature [7, 8] are a natural structural assumption for ranking distributions, allowing for particularly efficient processing of partial ranking queries.

## 1 Introduction

Rankings arise in a number of machine learning application settings such as preference analysis for movies and books [14] and political election analysis [6, 8]. In these applications, two central problems typically arise — (1) *representation*, how to efficiently build and represent a flexible statistical model, and (2) *reasoning*, how to efficiently use these statistical models to draw probabilistic inferences from observations. Both problems are challenging because of the fact that, as the number of items being ranked increases, the number of possible rankings increases factorially.

The key to efficient representations and reasoning is to identify exploitable problem structure, and to this end, there have been a number of smart structural assumptions proposed by the scientific community. These assumptions have typically been designed to reduce the number of necessary parameters of a model and have ranged from smoothness [10], to sparsity [11], to exponential family parameterizations [14].

We believe that these problems should be approached with the view that the two central challenges are intertwined — that *model structure should be chosen so that the most typical inference queries can be answered most efficiently.* So what are the most typical inference queries? In this paper, we assume that for ranking data, the most useful and typical inference queries take the form of *partial rankings*. For example, in election data, given a voter's top-$k$ favorite candidates in an election, we are interested in inferring his preferences over the remaining $n - k$ candidates.

Partial rankings are ubiquitous and come in myriad forms, from top-$k$ votes to approval ballots to rating data; Probability models over rankings that cannot efficiently handle partial ranking data therefore have limited applicability. Ranking datasets in fact are often predominantly composed of partial rankings rather than full rankings, and in addition, are often heterogenous, containing partial ranking data of mixed types.

In this paper, we contend that *the structural assumption of riffled independence is particularly well suited to answering probabilistic queries about partial rankings.* Riffled independence, which was introduced in recent literature by Huang et al. [7, 8], is a generalized notion of probabilistic independence for ranked data. Like graphical model factorizations based on conditional independence, riffled independence factorizations allow for flexible modeling of distributions over rankings which can be learned with low sample complexity. We show in particular, that when riffled independence assumptions are made about a prior distribution, partial ranking observations decompose in a way that allows for efficient conditioning. The main contributions of our work are as follows:

- When items satisfy the riffled independence relationship, we show that conditioning on partial rankings can be done efficiently, with running time linear in the number of model parameters.

- We show that, in a sense (which we formalize), it is impossible to efficiently condition on observations that do not take the form of partial rankings.
- We propose the first algorithm that is capable of efficiently estimating the structure and parameters of riffle independent models from heterogeneous collections of partially ranked data.
- We show results on real voting and preference data evidencing the effectiveness of our methods.

## 2 Riffled independence for rankings

A ranking, $\sigma$, of items in an item set $\Omega$ is a one-to-one mapping between $\Omega$ and a rank set $R = \{1, \ldots, n\}$ and is denoted using *vertical bar notation* as $\sigma^{-1}(1)|\sigma^{-1}(2)|\ldots|\sigma^{-1}(n)$. We say that $\sigma$ ranks item $i_1$ *before (or over)* item $i_2$ if the rank of $i_1$ is less than the rank of $i_2$. For example, $\Omega$ might be $\{Artichoke, Broccoli, Cherry, Date\}$ and the ranking $Artichoke|Broccoli|Cherry|Date$ encodes a preference of Artichoke over Broccoli which is in turn preferred over Cherry and so on. The collection of all possible rankings of item set $\Omega$ is denoted by $S_\Omega$ (or just $S_n$ when $\Omega$ is implicit).

Since there are $n!$ rankings of $n$ items, it is intractable to estimate or even explicitly represent arbitrary distributions on $S_n$ without making structural assumptions about the underlying distribution. While there are many possible simplifying assumptions that one can make, we focus on a recent approach [7, 8] in which the ranks of items are assumed to satisfy an intuitive generalized notion of probabilistic independence known as *riffled independence*. In this paper, we argue that riffled independence assumptions are particularly effective in settings where one would like to make queries taking the form of partial rankings. In the remainder of this section, we review riffled independence.

The riffled independence assumption posits that rankings over the item set $\Omega$ are generated by independently generating rankings of smaller disjoint item subsets (say, $A$ and $B$) which partition $\Omega$, and piecing together a full ranking by interleaving (or riffle shuffling) these smaller rankings together. For example, to rank our item set of foods, one might first rank the vegetables and fruits separately, then interleave the two subset rankings to form a full ranking. To formally define riffled independence, we use the notions of *relative rankings* and *interleavings*.

**Definition 1** (Relative ranking map). Given a ranking $\sigma \in S_\Omega$ and any subset $A \subset \Omega$, the *relative ranking of items in $A$*, $\phi_A(\sigma)$, is a ranking, $\pi \in S_A$, such that $\pi(i) < \pi(j)$ if and only if $\sigma(i) < \sigma(j)$.

**Definition 2** (Interleaving map). Given a ranking $\sigma \in S_\Omega$ and a partition of $\Omega$ into disjoint sets $A$ and $B$, the *interleaving of $A$ and $B$ in $\sigma$* (denoted, $\tau_{AB}(\sigma)$) is a (binary) mapping from the rank set $R = \{1, \ldots, n\}$ to $\{A, B\}$ indicating whether a rank in $\sigma$ is occupied by $A$ or $B$. As with rankings, we denote the interleaving of a ranking by its vertical bar notation: $[\tau_{AB}(\sigma)](1)|[\tau_{AB}(\sigma)](2)|\ldots|[\tau_{AB}(\sigma)](n)$.

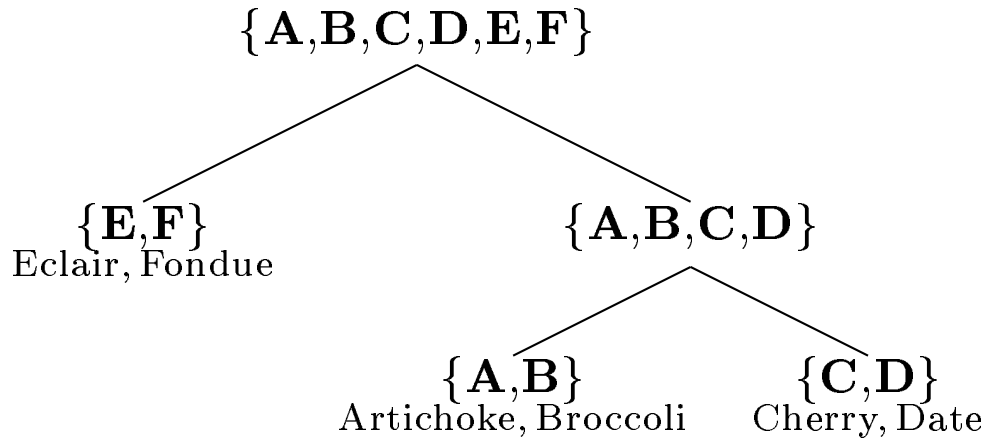

Figure 1: An example of a hierarchy over six food items.

**Example 3.** *Consider a partitioning of an item set $\Omega$ into vegetables $A = \{Artichoke, Broccoli\}$ and fruits $B = \{Cherry, Date\}$, and a ranking over these four items $\sigma = Artichoke|Date|Broccoli|Cherry$. In this case, the relative ranking of vegetables in $\sigma$ is $\phi_A(\sigma) = Artichoke|Broccoli$ and the relative ranking of fruits in $\sigma$ is $\phi_B(\sigma) = Date|Cherry$. The interleaving of vegetables and fruits in $\sigma$ is $\tau_{AB}(\sigma) = A|B|A|B$.*

**Definition 4** (Riffled Independence). Let $h$ be a distribution over $S_\Omega$ and consider a subset of items $A \subset \Omega$ and its complement $B$. The sets $A$ and $B$ are said to be *riffle independent* if $h$ decomposes (or factors) as:

$$h(\sigma) = m_{AB}(\tau_{AB}(\sigma)) \cdot f_A(\phi_A(\sigma)) \cdot g_B(\phi_B(\sigma)),$$

for distributions $m_{AB}$, $f_A$ and $g_B$, defined over interleavings and relative rankings of $A$ and $B$ respectively. We refer to $m_{AB}$ as the *interleaving distribution* and $f_A$ and $g_B$ as the *relative ranking distributions*.

Riffled independence has been found to approximately hold in a number of real datasets [9]. When such relationships can be identified in data, then instead of exhaustively representing all $n!$ ranking probabilities, one can represent just the factors $m_{AB}$, $f_A$ and $g_B$, which are distributions over smaller sets.

**Hierarchical riffle independent models.** The relative ranking factors $f_A$ and $g_B$ are themselves distributions over rankings. To further reduce the parameter space, it is natural to consider hierarchical decompositions of itemsets into nested collections of partitions (like hierarchical clustering). For example, Figure 2 shows a hierarchical decomposition where vegetables are riffle independent of fruits among the "healthy" foods, and these healthy foods are riffle independent of the subset $\{Eclair, Fondue\}$.

For simplicity, we restrict consideration to binary hierarchies, defined as tuples of the form $H = (H_A, H_B)$, where $H_A$ and $H_B$ are either (1) null, in which case $H$ is called a *leaf*, or (2) hierarchies over item sets $A$ and $B$ respectively. In this second case, $A$ and $B$ are assumed to be a nontrivial partition of the item set.

**Definition 5.** We say that a distribution $h$ factors riffle independently with respect to a hierarchy $H = (H_A, H_B)$ if item sets $A$ and $B$ are riffle independent with respect to $h$, and both $f_A$ and $g_B$ factor riffle independently with respect to subhierarchies $H_A$ and $H_B$, respectively.

Like Bayesian networks, these hierarchies represent families of distributions obeying a certain set of (riffled) independence constraints and can be parameterized locally. To draw from such a model, one generates full rankings recursively starting by drawing rankings of the leaf sets, then working up the tree, sequentially interleaving rankings until reaching the root. The parameters of these hierarchical models are simply the interleaving and relative ranking distributions at the internal nodes and leaves of the hierarchy, respectively.

By decomposing distributions over rankings into small pieces (like Bayesian networks have done for other distributions), these hierarchical models allow for better interpretability, efficient probabilistic representation, low sample complexity, efficient MAP optimization, and, as we show in this paper, efficient inference.

## 3 Decomposable observations

Given a prior distribution, $h$, over rankings and an observation $\mathcal{O}$, Bayes rule tells us that the posterior distribution, $h(\sigma|\mathcal{O})$, is proportional to $L(\mathcal{O}|\sigma) \cdot h(\sigma)$, where $L(\mathcal{O}|\sigma)$ is the likelihood function. This operation of *conditioning* $h$ on an observation $\mathcal{O}$ is typically computationally intractable since it requires multiplying two $n!$ dimensional functions, unless one can exploit structural decompositions of the problem. In this section, we describe a decomposition for a certain class of likelihood functions over the space of rankings in which the observations are 'factored' into simpler parts. When an observation $\mathcal{O}$ is decomposable in this way, we show that one can efficiently condition a riffle independent prior distribution on $\mathcal{O}$. For simplicity in this paper, we focus on *subset observations* whose likelihood functions encode membership with some subset of rankings in $S_n$.

**Definition 6** (Observations)**.** A *subset observation* $\mathcal{O}$ is an observation whose likelihood is proportional to the indicator function of some subset of $S_n$.

As a running example, we will consider the class of *first place observations* throughout the paper. The first place observation $\mathcal{O}$ ="Artichoke is ranked first", for example, is associated with the collection of rankings placing the item Artichoke in first place ($\mathcal{O} = \{\sigma : \sigma(\textit{Artichoke}) = 1\}$). In this paper, we are interested in computing $h(\sigma|\sigma \in \mathcal{O})$. In the first place scenario, we are given a voter's top choice and we would like to infer his preferences over the remaining candidates.

Given a partitioning of the item set $\Omega$ into two subsets $A$ and $B$, it is sometimes possible to *decompose* (or *factor*) a subset observation involving items in $\Omega$ into smaller subset observations involving $A$, $B$ and the interleavings of $A$ and $B$ independently. Such decompositions can often be exploited for efficient inference.

**Example 7.**

- *The first place observation $\mathcal{O}$ ="Artichoke is ranked first" can be decomposed into two independent observations — (1) an observation on the relative ranking of Vegetables ($\mathcal{O}_A$ ="Artichoke is ranked first among Vegetables"), and (2) an observation on the interleaving of Vegetables and Fruits, ($\mathcal{O}_{A,B}$ ="First place is occupied by a Vegetable"). To condition on $\mathcal{O}$, one updates the relative ranking distribution over Vegetables (A) by zeroing out rankings of vegetables which do not place Artichoke in first place, and updates the interleaving distribution by zeroing out interleavings which do not place a Vegetable in first place, then normalizes the resulting distributions.*
- *An example of a* nondecomposable *observation is the observation $\mathcal{O}$ ="Artichoke is in third place". To see that $\mathcal{O}$ does not decompose (with respect to Vegetables and Fruits), it is enough to notice that the interleaving of Vegetables and Fruits is not independent of the relative ranking of Vegetables. If, for example, an element $\sigma \in \mathcal{O}$ interleaves A (Vegetables) and B (Fruits) as $\tau_{AB}(\sigma) = A|B|A|B$, then since $\sigma(\textit{Artichoke}) = 3$, the relative ranking of Vegetables is constrained to be $\phi_A(\sigma) = \textit{Broccoli}|\textit{Artichoke}$. Since the interleavings and relative rankings are not independent, we see that $\mathcal{O}$ cannot be decomposable.*

Formally, we use riffle independent factorizations to define decomposability with respect to a hierarchy $H$ of the item set.

**Definition 8** (Decomposability)**.** Given a hierarchy $H$ over the item set, a subset observation $\mathcal{O}$ *decomposes* with respect to $H$ if its likelihood function $L(\mathcal{O}|\sigma)$ factors riffle independently with respect to $H$.

When subset observations and the prior decompose according to the same hierarchy, we can show (as in Example 7) that the posterior also decomposes.

**Proposition 9.** *Let $H$ be a hierarchy over the item set. Given a prior distribution $h$ and an observation $\mathcal{O}$ which both decompose with respect to $H$, the posterior distribution $h(\sigma|\mathcal{O})$ also factors riffle independently with respect to $H$.*

In fact, we can show that when the prior and observation both decompose with respect to the same hierarchy, *inference operations can always be performed in time linear in the number of model parameters.*

## 4 Complete decomposability

The condition of Proposition 9, that the prior and observation must decompose with respect to *exactly* the same hierarchy is a sufficient one for efficient inference, but it might at first glance seem so restrictive as to render the proposition useless in practice. To overcome this limitation of "hierarchy specific" decomposability, we explore a special family of observations (which we call *completely decomposable*) for which the property of decomposability does not depend specifically on a particular hierarchy, implying in particular that for these observations, efficient inference is *always* possible (as long as efficient representation of the prior distribution is also possible).

To illustrate how an observation can decompose with respect to multiple hierarchies, consider again the first place observation $\mathcal{O}$ ="Artichoke is ranked first". We argued in Example 7 that $\mathcal{O}$ is a decomposable observation. Notice however that decomposability for this particular observation *does not* depend on how the items are partitioned by the hierarchy. Specifically, if instead of Vegetables and Fruits, the sets $A = \{Artichoke, Cherry\}$ and $B = \{Broccoli, Date\}$ are riffle independent, a similar decomposition of $\mathcal{O}$ would continue to hold, with $\mathcal{O}$ decomposing as (1) an observation on the relative ranking of items in $A$ ("Artichoke is first among items in $A$"), and (2) an observation on the interleaving of $A$ and $B$ ("First place is occupied by some element of $A$").

To formally capture this notion that an observation can decompose with respect to *arbitrary* underlying hierarchies, we define *complete decomposability*:

**Definition 10** (Complete decomposability)**.** We say that an observation $\mathcal{O}$ is *completely decomposable* if it decomposes with respect to *every* possible hierarchy over the item set $\Omega$.

The property of complete decomposability is a *guarantee* for an observation $\mathcal{O}$, that one can always exploit any available factorized structure of the prior distribution in order to efficiently condition on $\mathcal{O}$.

**Proposition 11.** *Given a prior $h$ which factorizes with respect to a hierarchy $H$, and a completely decomposable observation $\mathcal{O}$, the posterior $h(\sigma|\mathcal{O})$ also decomposes with respect to $H$ and can be computed in time linear in the number of model parameters of $h$.*

Given the stringent conditions in Definition 10, it is not obvious that nontrivial completely decomposable observations even exist. Nonetheless, there do exist nontrivial examples (such as the first place observations), and in the next section, we exhibit a rich and general class of completely decomposable observations.

## 5 Complete decomposability of partial ranking observations

In this section we discuss the mathematical problem of identifying all completely decomposable observations. Our main contribution in this section is to show that *completely decomposable observations correspond exactly to partial rankings of the item set.*

**Partial rankings.** We begin our discussion by introducing *partial rankings*, which allow for items to be tied with respect to a ranking $\sigma$ by 'dropping' verticals from the vertical bar representation of $\sigma$.

**Definition 12** (Partial ranking observation)**.** Let $\Omega_1$, $\Omega_2, \ldots, \Omega_k$ be an ordered collection of subsets which partition $\Omega$ (i.e., $\cup_i \Omega_i = \Omega$ and $\Omega_i \cap \Omega_j = \emptyset$ if $i \neq j$). The *partial ranking observation*[1] corresponding to this partition is the collection of rankings which rank items in $\Omega_i$ before items in $\Omega_j$ if $i < j$. We denote this partial ranking as $\Omega_1|\Omega_2|\ldots|\Omega_k$ and say that it has type $\gamma = (|\Omega_1|, |\Omega_2|, \ldots, |\Omega_k|)$.

Given the type $\gamma$ and any full ranking $\pi \in S_\Omega$, there is only one partial ranking of type $\gamma$ containing $\pi$, thus we will also equivalently denote the partial ranking $\Omega_1|\Omega_2|\ldots|\Omega_k$ as $S_\gamma \pi$, where $\pi$ is any element of $\Omega_1|\Omega_2|\ldots|\Omega_k$.

The space of partial rankings as defined above captures a rich and natural class of observations. In particular, partial rankings encompass a number of commonly occurring special cases, which have traditionally been modeled in isolation, but in our work (as well as recent works such as [14, 13]) can be used in a unified setting.

**Example 13.** *Partial ranking observations include:*

- (First place, or Top-1 observations): *First place observations correspond to partial rankings of type $\gamma = (1, n-1)$. The observation that Artichoke is ranked first can be written as Artichoke|Broccoli, Cherry, Date.*
- (Top-$k$ observations): *Top-$k$ observations are partial rankings with type $\gamma = (1, \ldots, 1, n-k)$. These generalize the first place observations by specifying the items mapping to the first $k$ ranks, leaving all $n-k$ remaining items implicitly ranked behind.*
- (Desired/less desired dichotomy): *Partial rankings of type $\gamma = (k, n-k)$ correspond to a subset of $k$ items being preferred or desired over the remaining subset of $n-k$ items. For example, partial rankings of type $(k, n-k)$ might arise in* approval voting *in which voters mark the subset of "approved" candidates, implicitly indicating disapproval of the remaining $n-k$ candidates.*

**Single layer decomposition.** To show how partial rankings observations decompose, we will exhibit an explicit factorization with respect to a hierarchy $H$ over items. For simplicity, we begin by considering the single layer case, in which the items are partitioned into two leaf sets $A$ and $B$. Our factorization depends on the following notions of *consistency* of relative rankings and interleavings with a partial ranking.

[1]As in [1], we note that "The term *partial ranking* used here should not be confused with two other standard objects: (1) Partial order, namely, a reflexive, transitive antisymmetric binary relation; and (2) A ranking of a subset of $\Omega$ . In search engines, for example, although only the top-$k$ elements of $\Omega$ are returned, the remaining $n-k$ are implicitly assumed to be ranked behind."

**Definition 14** (Restriction consistency). Given a partial ranking $S_\gamma\pi = \Omega_1|\Omega_2|\ldots|\Omega_k$ and any subset $A \subset \Omega$, we define the *restriction* of $S_\gamma\pi$ to $A$ as the partial ranking on items in $A$ obtained by intersecting each $\Omega_i$ with $A$. Hence the restriction of $S_\gamma\pi$ to $A$ is:

$$[S_\gamma\pi]_A = \Omega_1 \cap A|\Omega_2 \cap A|\ldots|\Omega_k \cap A.$$

Given a ranking, $\sigma_A$ of items in $A$, we say that $\sigma_A$ is *consistent* with the partial ranking $S_\gamma\pi$ if $\sigma_A$ is a member of $[S_\gamma\pi]_A$.

**Definition 15** (Interleaving consistency). Given an interleaving $\tau_{AB}$ of two sets $A, B$ which partition $\Omega$, we say that $\tau_{AB}$ is *consistent* with a partial ranking $S_\gamma\pi = \Omega_1|\ldots|\Omega_k$ (with type $\gamma$) if the first $\gamma_1$ entries of $\tau_{AB}$ contain the same number of As and Bs as $\Omega_1$, and the second $\gamma_2$ entries of $\tau_{AB}$ contain the same number of As and Bs as $\Omega_2$, and so on. Given a partial ranking $S_\gamma\pi$, we denote the collection of consistent interleavings as $[S_\gamma\pi]_{AB}$.

For example, consider the partial ranking $S_\gamma\pi = Artichoke, Cherry|Broccoli, Date$, which places a single vegetable and a single fruit in the first two ranks, and a single vegetable and a single fruit in the last two ranks. Alternatively, $S_\gamma\pi$ partially specifies an interleaving $AB|AB$. The full interleavings $A|B|B|A$ and $B|A|B|A$ are consistent with $S_\gamma\pi$ (by dropping vertical lines) while $A|A|B|B$ is *not* consistent (since it places two vegetables in the first two ranks).

Using the notions of consistency with a partial ranking, we show that partial ranking observations are decomposable with respect to any binary partitioning (i.e., single layer hierarchy) of the item set.

**Proposition 16** (Single layer hierarchy). *For any partial ranking observation $S_\gamma\pi$ and any binary partitioning of the item set $(A, B)$, the indicator function of $S_\gamma\pi$, $\delta_{S_\gamma\pi}$, factors riffle independently as:*

$$\delta_{S_\gamma\pi}(\sigma) = m_{AB}(\tau_{AB}(\sigma)) \cdot f_A(\phi_A(\sigma)) \cdot g_B(\phi_B(\sigma)), \quad (5.1)$$

*where the factors $m_{AB}$, $f_A$ and $g_B$ are the indicator functions for consistent interleavings and relative rankings, $[S_\gamma\pi]_{AB}$, $[S_\gamma\pi]_A$ and $[S_\gamma\pi]_B$, respectively.*

**General hierarchical decomposition.** The single layer decomposition of Proposition 16 can be turned into a recursive decomposition for partial ranking observations over arbitrary binary hierarchies, which establishes our main result. In particular, given a partial ranking $S_\gamma\pi$ and a prior distribution which factorizes according to a hierarchy $H$, we first condition the topmost interleaving distribution by zeroing out all parameters corresponding to interleavings which are not consistent with $S_\gamma\pi$, and normalizing the distribution.

We then need to condition the subhierarchies $H_A$ and $H_B$ on relative rankings of $A$ and $B$ which are consistent with $S_\gamma\pi$, respectively. Since these consistent sets, $[S_\gamma\pi]_A$ and $[S_\gamma\pi]_B$, are partial rankings themselves, the same algorithm for conditioning on a partial ranking can be applied recursively to each of the subhierarchies $H_A$ and $H_B$.

**Theorem 17.** *Every partial ranking is completely decomposable.*

See Algorithm 1 for details on our recursive conditioning algorithm. As a consequence of Theorem 17 and Proposition 11, conditioning on partial ranking observations can be performed in linear time with respect to the number of model parameters.

It is interesting to consider what completely decomposable observations exist beyond partial rankings. One of our main contributions is to show that there are no such observations.

**Theorem 18** (Converse of Theorem 17). Every *completely decomposable observation takes the form of a partial ranking.*

Together, Theorems 17 and 18 form a significant insight into the nature of rankings, showing that the notions of partial rankings and complete decomposability *exactly* coincide. In fact, our result shows that it is even possible to define partial rankings via complete decomposability!

As a practical matter, our results show that there is no algorithm based on simple multiplicative updates to the parameters which can exactly condition on observations which do *not* take the form of partial rankings. If one is interested in conditioning on such observations, Theorem 18 suggests that a slower or approximate inference approach might be necessary.

# 6 Model estimation from partially ranked data

In this section we use the efficient inference algorithm proposed in Section 5 for estimating a riffle independent model from partially ranked data. Because estimating a model using partially ranked data is typically considered to be more difficult than estimating one using only full rankings, a common practice (see for example [8]) has been to simply ignore the partial rankings in a dataset. The ability of a method to incorporate *all* of the available data however, can lead to significantly improved model accuracy as well as wider applicability of that method. In this section, *we propose the first efficient method for estimating the structure and parameters of a hierarchical riffle independent model from heterogeneous datasets consisting of arbitrary partial ranking types.* Central to our approach is the idea that given someone's partial preferences, **function** PRCONDITION (Prior $h_{prior}$, Hierarchy $H$, Observation $S_\gamma\pi = \Omega_1|\Omega_2|\dots|\Omega_k$)

**if** *isLeaf(H)* **then**
  **forall** $\sigma$ **do**
    $$h_{post}(\sigma) \leftarrow \begin{cases} h_{prior}(\sigma) & \text{if } \sigma \in S_\gamma\pi \\ 0 & \text{otherwise} \end{cases};$$
  NORMALIZE ($h_{post}$) ;
  **return** $h_{post}$;
**else**
  **forall** $\tau$ **do**
    $$m_{post}(\tau) \leftarrow \begin{cases} m_{prior}(\tau) & \text{if } \tau \in [S_\gamma\pi]_{AB} \\ 0 & \text{otherwise} \end{cases};$$
  NORMALIZE ($m_{post}$) ;
  $f(\phi_A) \leftarrow$PRCONDITION ($f_{prior}, H_A, [S_\gamma\pi]_A$) ;
  $g(\phi_B) \leftarrow$PRCONDITION ($g_{prior}, H_B, [S_\gamma\pi]_B$) ;
  **return** $m_{post}, f_{post}, g_{post}$;

**Algorithm 1**: Pseudocode for PRCONDITION, an algorithm for recursively conditioning a hierarchical riffle independent prior distribution on partial ranking observations. See Definitions 14 and 15 for $[S_\gamma\sigma]_A$, $[S_\gamma\sigma]_B$, and $[S_\gamma\sigma]_{AB}$. The runtime of PRCONDITION is linear in the number of model parameters.

we can use the efficient algorithms developed in the previous section to infer his full preferences and consequently apply previously proposed algorithms which are designed to work with full rankings.

**Censoring interpretations of partial rankings.** The model estimation problem for full rankings is stated as follows. Given i.i.d. training examples $\sigma^{(1)}, \dots, \sigma^{(m)}$ (consisting of full rankings) drawn from a hierarchical riffle independent distribution $h$, recover the structure and parameters of $h$.

In the partial ranking setting, we again assume i.i.d. draws, but that each training example $\sigma^{(i)}$ undergoes a censoring process producing a partial ranking consistent with $\sigma^{(i)}$. For example, censoring might only allow for the ranking of the top-$k$ items of $\sigma^{(i)}$ to be observed. While we allow for arbitrary types of partial rankings to arise via censoring, we make a common assumption that the partial ranking type resulting from censoring $\sigma^{(i)}$ does not depend on $\sigma^{(i)}$ itself.

**Algorithm.** We treat the model estimation from partial rankings problem as a missing data problem. As with many such problems, if we could determine the full ranking corresponding to each observation in the data, then we could apply algorithms which work in the completely observed data setting. Since full rankings are not given, we utilize an Expectation-Maximization (EM) approach in which we use inference to compute a posterior distribution over full rankings given the observed partial ranking. In our case, we then apply the algorithms from [8, 9] which were designed to estimate the hierarchical structure of a model and its parameters from a dataset of full rankings.

Given an initial model $h$, our EM-based approach alternates between the following two steps until convergence is achieved.

- **(E-step)**: For each partial ranking, $S_\gamma\pi$, in the training examples, we use inference to compute a posterior distribution over the full ranking $\sigma$ that could have generated $S_\gamma\pi$ via censoring, $h(\sigma|\mathcal{O} = S_\gamma\pi)$. Since the observations take the form of partial rankings, we use the efficient algorithms in Section 5 to perform the E-step.
- **(M-step)**: In the M-step, one maximizes the expected log-likelihood of the training data with respect to the model. When the hierarchical structure of the model has been provided, or is known beforehand, our M-step can be performed using standard methods for optimizing parameters. When the structure is *unknown*, we use a *structural EM* approach, which is analogous to methods from the graphical models literature for structure learning from incomplete data [4, 5].

  Unfortunately, the (riffled independence) structure learning algorithm of [8] is unable to directly use the posterior distributions computed from the E-step. Instead, observing that sampling from riffle independent models can be done efficiently and exactly (as opposed to, for example, MCMC methods), we simply sample full rankings from the posterior distributions computed in the E-step and pass these full rankings into the structure learning algorithm of [8]. The number of samples that are necessary, instead of scaling factorially, scales according to the number of samples required to detect riffled independence (which under mild assumptions is polynomial in $n$, [8]).

**Related work.** There are several recent works to model partial rankings. Busse et al. [2] learned finite mixtures of Mallows models from top-$k$ data (also using an EM approach). Lebanon and Mao [14] developed a nonparametric model based (also) on Mallows models which can handle arbitrary types of partial rankings. In both settings, a central problem is to marginalize a Mallows model over all full rankings which are consistent with a particular partial ranking. To do so efficiently, both papers rely on the fact (first shown in [3]) that this marginalization step can be performed in closed form. This closed form equation of [3], however, can be seen as a very special case of our setting since Mallows models can always be shown to factor riffle independently according to a chain structure.[2] Moreover, instead of resorting to the more complicated mathematics based on inversion combinatorics, our theory of complete decomposability offers a simple conceptual way to understand why Mallows models can be conditioned efficiently on partial ranking observations.

[2]A chain structure is a hierarchy in which only a single item is partitioned out at each level of the hierarchy.

## 7 Experiments

We demonstrate our algorithms on simulated data as well as real datasets taken from two different domains.

The *Meath dataset* [6] is taken from a 2002 Irish Parliament election with over 60,000 top-$k$ rankings of 14 candidates. Figure 2(a) plots, for each $k \in \{1, \ldots, 14\}$, the number of ballots in the Meath data of length $k$. In particular, note that the vast majority of ballots in the dataset consist of partial rather than full rankings. We can run inference on over 5000 top-$k$ examples for the Meath data in 10 seconds on a dual 3.0 GHz Pentium machine with an unoptimized Python implementation. Using 'brute force' inference, we estimate that the same job would require roughly one hundred years.

We extracted a second dataset from a database of *searchtrails* collected by [15], in which browsing sessions of roughly 2000 users were logged during 2008-2009. In many cases, users are unlikely to read articles about the same story twice, and so it is often possible to think of the order in which a user reads through a collection of articles as a top-$k$ ranking over articles concerning a particular story/topic. The ability to model visit orderings would allow us to make long term predictions about user browsing behavior, or even recommend 'curriculums' over articles for users. We ran our algorithms on roughly 300 visit orderings for the eight most popular posts from `www.huffingtonpost.com` concerning 'Sarah Palin', a popular subject during the 2008 U.S. presidential election. Since no user visited every article, there are no full rankings in the data and thus the method of 'ignoring' partial rankings does not work.

**Structure discovery with EM** In all experiments, we initialize distributions to be uniform, and do not use random restarts. Our experiments have led to several observations about using EM for learning with partial rankings. First, we observe that typical runs converge to a fixed structure quickly, with no more than three EM iterations. Figure 2(b) shows the progress of EM on the Sarah Palin data, whose structure converges by the third iteration. As expected, the log-likelihood increases at each iteration, and we remark that the structure becomes more interpretable — for example, the leaf set $\{0, 2, 3\}$ corresponds to the three posts about Palin's wardrobe before the election, while the posts from the leaf set $\{1, 4, 6\}$ were related to verbal gaffes made by Palin during the campaign.

Secondly, the number of EM iterations required to reach convergence in log-likelihood depends on the types of partial rankings observed. We ran our algorithm on subsets of the Meath dataset, each time training on $m = 2000$ rankings all with length larger than some fixed $k$. Figure 2(c) shows the number of iterations required for convergence as a function of $k$ (with 20 bootstrap trials for each $k$). We observe fastest convergence for datasets consisting of almost-full rankings and slowest convergence for those consisting of almost-empty rankings, with almost 25 iterations necessary if one trains using rankings of all types.

**The value of partial rankings.** We now show that using partial rankings in addition to full rankings allows us to achieve better density estimates. We first learned models from synthetic data drawn from a hierarchy, training using 343 full rankings plus varying numbers of partial ranking examples (ranging between 0-64,000). We repeat each setting with 20 bootstrap trials, and for evaluation, we compute the log-likelihood of a testset with 5000 examples. For speed, we learn a structure $H$ only once and fix $H$ to learn parameters for each trial. Figure 2(d), which plots the test log-likelihood as a function of the number of partial rankings made available to the training set, shows that we are indeed able to learn more accurate distributions as more and more data are made available.

**Comparing to a nonparametric model.** Comparing the performance of riffle independent models to other approaches was not previously possible since [8] could not handle partial rankings. Using our methods, we compare riffle independent models with the state-of-the-art nonparametric Lebanon-Mao (LM08) estimator of [14] on the same data (setting their regularization parameter to be $C$ =1,2,5, or 10 via a validation set). Figure 2(d) shows (naturally) that when the data are drawn synthetically from a riffle independent model, then our EM method significantly outperforms the LM08 estimator.

For the Meath data, which is only approximately riffle independent, we trained on subsets of size 5,000 and 25,000 (testing on remaining data). For each subset, we evaluated our EM algorithm for learning a riffle independent model against the LM08 estimator when (1) using only full ranking data, and (2) using all data. As before, both methods do better when partial rankings are made available.

For the smaller training set, the riffle independent model performs as well or better than the LM'08 estimator. For the larger training set of 25,000, we see that the nonparametric method starts to perform slightly better on average, the advantage of a nonparametric model being that it is guaranteed to be consistent, converging to the correct model given enough data. The advantage of riffle independent models, however, is that they are simple, interpretable, and can highlight global structures hidden within the data.

## 8 Conclusion

In probabilistic reasoning problems, it is often the case that certain data types suggest certain distribution representations. For example, sparse dependency

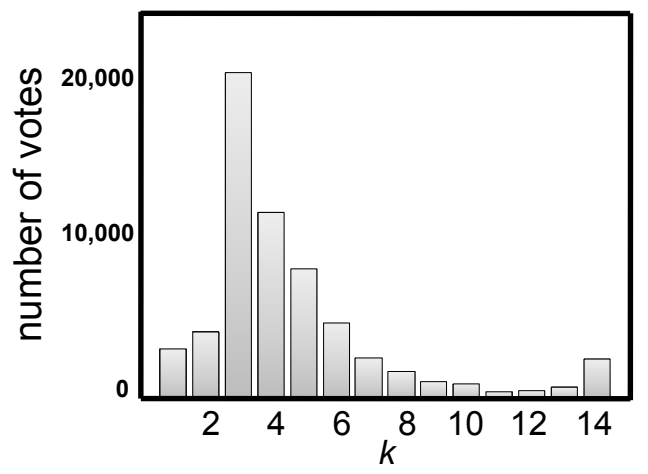

(a) Histogram of top-$k$ ballot lengths in the Irish election data. Over half of voters provide only their top-3 or top-4 choices.

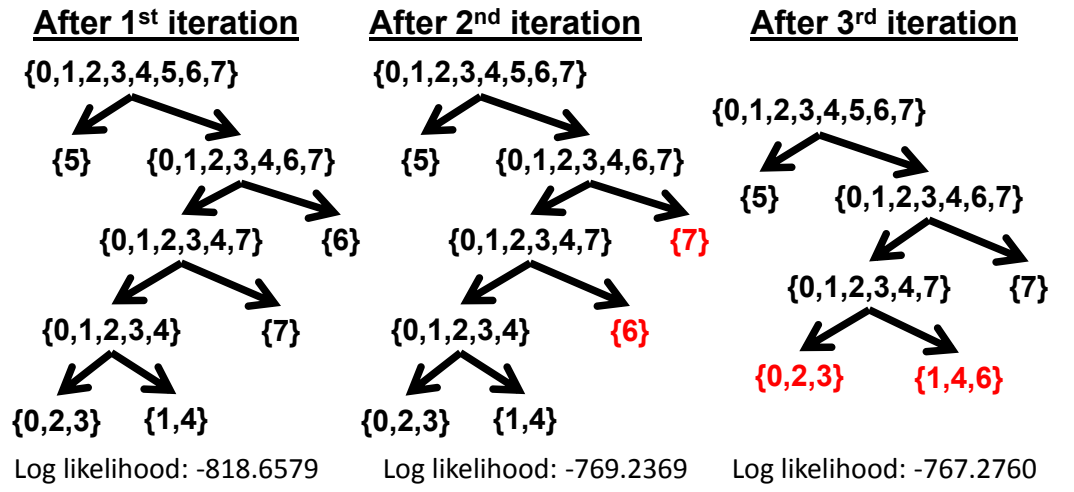

(b) Iterations of Structure EM for the Sarah Palin data with structural changes at each iteration highlighted in red. This figure is best viewed in color.

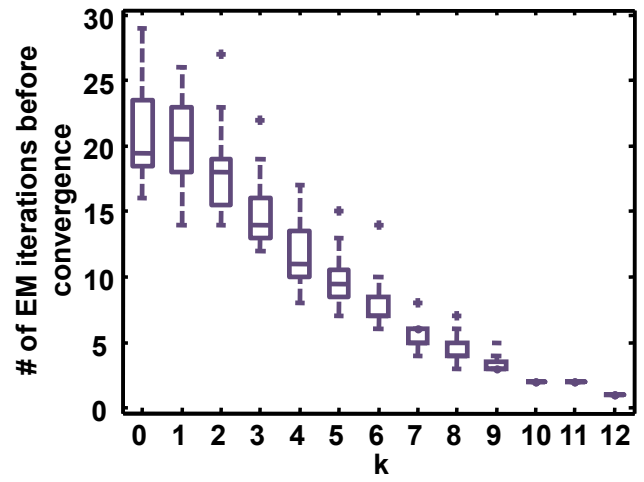

(c) # of EM iterations required for convergence if the training set only contains rankings of length longer than $k$.

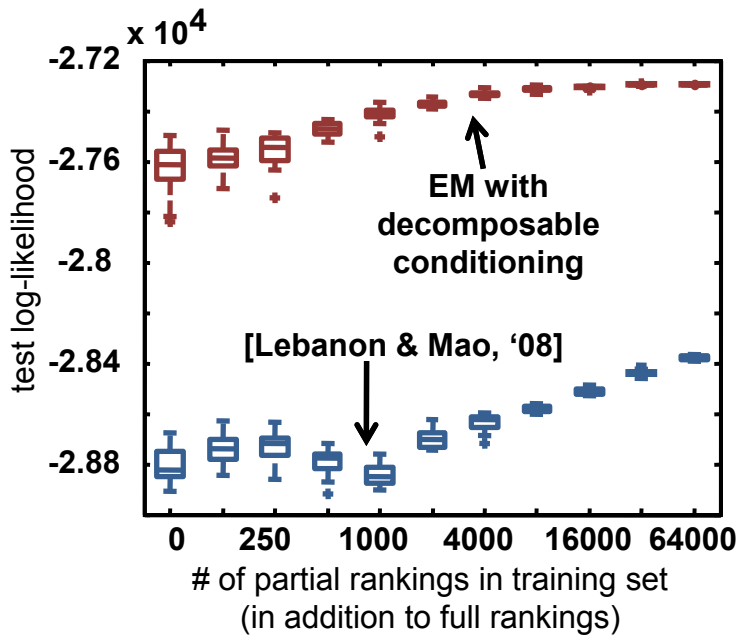

(d) Density estimation from synthetic data. We plot test loglikelihood when learning from 343 full rankings and between 0 and 64,000 additional partial rankings.

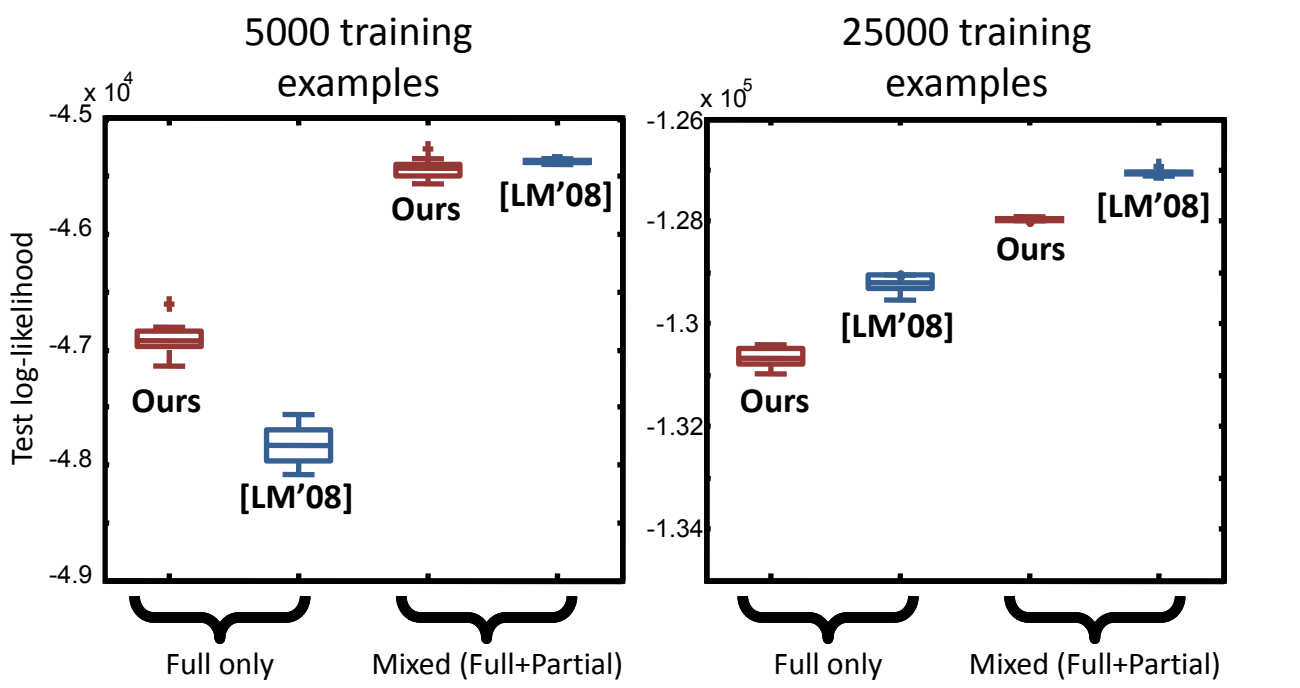

(e) Density estimation from small (5000 examples) and large subsets (25000 examples) of the Meath data. We compare our method against [14] training (1) on all available data and (2) on the subset of full rankings.

Figure 2: Experimental results

structure in the data often suggests a Markov random field (or other graphical model) representation [4, 5]. For low-order permutation observations (depending on only a few items at a time), recent work ([10, 12]) has shown that a Fourier domain representation is appropriate. Our work shows, on the other hand, that when the observed data takes the form of partial rankings, then hierarchical riffle independent models are a natural representation.

As with conjugate priors, we showed that a riffle independent model is guaranteed to retain its factorization structure after conditioning on a partial ranking (which can be performed in linear time). Most surprisingly, our work shows that observations which do not take the form of partial rankings are not amenable to simple multiplicative update based conditioning algorithms. Finally, we showed that it is possible to learn hierarchical riffle independent models from partially ranked data, significantly extending the applicability of previous work.

### Acknowledgements

This work is supported in part by ONR under MURI N000140710747, and ARO under MURI W911NF0810242. C. Guestrin was funded in part by NSF Career IIS-064422. We thank E. Horvitz, R. White, and D. Liebling for discussions. Much of this research was conducted while J. Huang was visiting Microsoft Research Redmond.